# Eigen Values Features for the Classification of Brain Signals corresponding to 2D and 3D Educational Contents

Saeed Bamatraf[1], Muhammad Hussain[1], Emad-ul-Haq Qazi[1] and Hatim Aboalsamh[1]

[1]Visual Computing Lab, Department of Computer Science, College of Computer and Information Sciences, King Saud University, Riyadh, Saudi Arabia.

**Abstract**

*In this paper, we have proposed a brain signal classification method, which uses eigenvalues of the covariance matrix as features to classify images (topomaps) created from the brain signals. The signals are recorded during the answering of 2D and 3D questions. The system is used to classify the correct and incorrect answers for both 2D and 3D questions. Using the classification technique, the impacts of 2D and 3D multimedia educational contents on learning, memory retention and recall will be compared. The subjects learn similar 2D and 3D educational contents. Afterwards, subjects are asked 20 multiple-choice questions (MCQs) associated with the contents after thirty minutes (Short-Term Memory) and two months (Long-Term Memory). Eigenvalues features extracted from topomaps images are given to K-Nearest Neighbor (KNN) and Support Vector Machine (SVM) classifiers, in order to identify the states of the brain related to incorrect and correct answers. Excellent accuracies obtained by both classifiers and by applying statistical analysis on the results, no significant difference is indicated between 2D and 3D multimedia educational contents on learning, memory retention and recall in both STM and LTM.*

*Keywords*--- **Eigen values; EEG; Short-Term Memory; Long-Term Memory; Topomaps; KNN; SVM;**

## 1. Introduction

Multimedia content has played a vital role for education sector to educate the individuals of every domain from children to old people because it makes the concepts easy to understand. The multimedia principle by Mayer stated that people learned more deeply from words and pictures than from words alone [1]. Present multimedia related to educational content is in use of existing resources that are commonly 2D in nature, i.e., 2D displays and cameras. However, on the other side, 3D educational content and hardware become available widespread. Therefore, the utilization of 3D multimedia educational content will be more than the 2D in considerably shorter period of time. Therefore the educational contents can be shown to the students either in 2D or 3D format. In the 2D, the content is shown traditionally in a two-dimensional artistic space as presented in most of the educational organizations. Before the advent of 3D devices, 3D was previously shown as 2D. Nowadays, 3D can be shown as 3D with the help of 3D devices. These rapid developments within the field of multimedia technology have raised the question that is 3D multimedia educational content useful than 2D in the context of learning, information retention and recall?

Electroencephalography (EEG) is a painless and non-invasive brain mapping procedure, which can be utilized to record directly various states within the brain, and it can also be used to evaluate the memory recall and learning. Due to its painless and non-invasive property, it is currently being widely utilized in memory research to examine the cognitive processes of brain, i.e., emotion, language, attention, memory and perception in children and normal adults [2].

To answer the question regarding the effects of 3D and 2D educational contents on learning and memory recall, Saeed et al. [3] employed a pattern recognition system for the prediction of memory states as true or false memories using EEG brain signals. In this system, the features are extracted from the EEG signals corresponding to correct (true memory) and incorrect (false memory) answers. First, EEG signals are converted into topomaps, and redundant topomaps are removed using city-block distance, then first order statistics are used for feature extraction. In this paper, Eigen values based feature extraction technique has been proposed for this system and the main contribution of this paper is the very low number of features used by the system.

For image classification tasks, feature extraction is one of the most important steps because if discrimination of features is high then accuracy rate of the classifier is also high. There are many techniques in the literature for feature extraction. In this paper, we select the eigenvalues of the covariance matrix.

The eigenvector with the largest eigenvalue is the direction along which the data set has the maximum variance. The covariance matrix defines both the spread (variance), and the orientation (covariance) of our data. Therefore, if we would like to represent the covariance matrix with a vector and its magnitude, we should simply try to find the vector that points into the direction of the largest spread of the data, and whose magnitude equals the spread (variance) in this direction.

The idea of using eigenvalues of the covariance matrix as features is motivated from the characteristics of nuclear norm [4]. Each question is represented as a matrix and we extract eigenvalues of this matrix as features. These eigenvalues represent the variation along the variance of EEG signals and the highest variance along principle components.

The mean prediction accuracy of the system with the proposed eigenvalues based on the feature extraction technique is 98% for 2D and 99% for 3D in case of STM, and 98.5% for 2D and 99.5% for 3D in LTM case. Statistical analysis of the results indicates that there is no significant difference between 2D and 3D educational content on learning and memory recall for STM and LTM.

This paper is organized as follows: Section 2 gives some related works. Section 3 presents the proposed methodology. Section 4 presents the results and discussion, and section 5 concludes the paper.

## 2. Related Work:

The differences between 2D and 3D contents have been investigated during different tasks such as spatial cognition tasks [5-7], spatial visualization skills [8], in-depth understanding of PC hardware [9], movie content experience [10], knowledge acquisition [11, 12], educational learning processes [13]. However, these studies deduce the differences based on subjective responses without using and analyzing EEG brain signals. Furthermore, none of the researchers explicitly studied the effects of 2D and 3D educational contents on learning and memory recall. The researchers used the subjective approach based on the statistical analysis of the answers of the questions to assess the differences between 2D and 3D content; however, the subjective approach cannot give direct insight into brain states like EEG.

Many researchers implemented EEG technology to study brain activation during different tasks like cognitive tasks [14], playing video games on large screens [15, 16]. Some researchers studied the EEG brain signals to locate brain regions and the components responsible for memory functions [17-20]. Also some studies have been done recently on attention, learning, and memory using EEG brain activations [21-24]. To the best of our knowledge, there is no other research than [3, 25, 26] focused on studying the impact of 2D and 3D educational contents on learning and memory recall using direct brain behavior through EEG technology.

## 3. Methodology

In this section, we give an overview of the approach that we adopted to analyze the effects of 2D and 3D educational contents on learning and memory retention and recall. Then we present the detail of a new feature extraction technique to represent the brain states using EEG signals, which is the main contribution of the paper.

We developed two pattern recognition systems, one for 2D and the other for 3D, for assessing the brain states while true and false answers of MCQs i.e. true and false memories. An overview of such a system is shown in Fig. 1. Feature extraction is an important component of the system, and discriminative features are needed to represent the topomaps.

First, subjects are selected to participate in the learning and memory recalls tasks. EEG signals are recorded from the subjects during the tasks. These signals are preprocessed to remove the artifacts and noisy signals. Topographic maps (topomaps) are then created from the clean signals, and not like previous researches [3, 25, 26], we used all the topomaps without any selection of some of them. Eigen values of the covariance matrix are extracted as features from all topomaps. There is no need for feature selection because the highest eigenvalues clearly represent the questions and there is no need to reduce the dimensionality of feature spaces because the number of features used is very small. Using the classification model, we classify incorrect and correct answers of the questions and then highlight the impacts of 2D and 3D multimedia educational contents on learning, memory retention and recall. More details about the steps and how topomaps look like are already exist in our previous research [3]. We give a brief overview in the following sections.

### 3.1. Data Collection

A total number of sixty-six volunteers participated in the experiments, and their ages were in the range of eighteen to thirty years. The volunteers were not suffering from any form of neurological disorders and having normal or close to normal eye sight. Two groups were formed depending upon the level of knowledge and age of subjects for 2D and 3D educational content. Ethics Coordination Committee of Universiti Teknologi PETORNAS (UTP), Tronoh, Malaysia has approved this research work.

The experiments which are carried out in this research work comprised of two tasks, i.e., learning of educational contents and information recall from memory. In the learning task, the participants viewed 2D or 3D learning contents depending upon their group for the time spans of eight to ten minutes. In the phase of memory recall, the retention period was of thirty minutes for STM and of two months for LTM. In this process, twenty MCQs were inquired of the participants with 30 seconds time limit to answer each MCQ, and each MCQ had four choices for answers. The same MCQs were asked to participants of both the groups; that is, 2D and 3D. While performing the recall experiment, EEG signals were recorded for study.

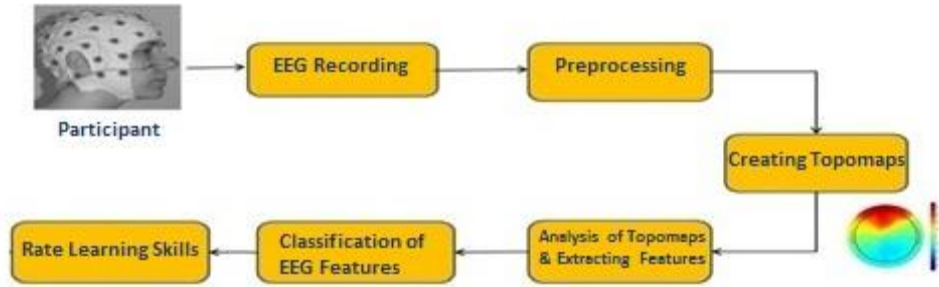

Figure 1: The architecture of the system

### 3.2 EEG Recording

The recording of EEG signals was done in the period of 30 seconds, which started when the participants were shown the questions and ended when they chose answers from the four choices for the answer.

At the point when question showed up on the screen, EEG signal recording relating to that particular question began, and it proceeds until the response is given by the subject by selecting one option out of four as a correct answer. A period of thirty seconds was given to each question. The starting time of displaying the question on the screen and the ending time when the answer is given by the subject by pressing the button are to be saved in an event file. This file will be used later to get the portion of relevant EEG signal relating to the question. Data concerning to EEG was recorded through applying the sampling rate of 250 samples per second using 128-channels Hydro Cel Geodesic Net, USA [27].

### 3.3 Preprocessing

The noise in the form of artifacts such as eye blinking and so forth is present in the recorded EEG data, which adversely affects the performance of feature extraction and eventually the prediction accuracy. It is utmost important to remove such noise for the overall performance of the system. In preprocessing phase, such noise is removed. The noise is present in various forms in the recorded EEG data, which includes the EEG activity that is not the result of response of stimuli; noise due to the variability in ERP components is a result of neural and cognitive activity variations; another common source of noise is the presence of bioelectric activities like movement of eyes, blinking, movement of muscles, and so forth; and the final source of noise is due to the electric equipment like display devices and so forth.

Raw EEG data was filtered by applying the band pass filter (1-48 Hz). The artifacts were identified. In the next step, data was exported into .mat files format (Matlab) by utilizing the Netstation software of EGI. Visual examination, and Gratton and Coles method [28] were utilized to remove the ocular artifacts in the recorded data.

### 3.4 Topomaps Creation

Topographical maps (topomaps) are related with voltages of channels in EEG signals. They are represented as images. Consequently, different image analysis and processing techniques can be utilized to get the desired features from topomaps images.

The EEG signals which are recorded from each subject while answering all the twenty questions is saved in a Matlab raw file (.mat) and the time points at which the subject starts answering each question until finishes answering are saved in an event file corresponding to each subject.

Using .mat file together with event files corresponding to each question, we created the topomaps corresponding to the EEG recording of the specific question of the recall task. We used EEGLAB toolbox [29] in Matlab at this step. The number of topomaps in each question is determined by the time when subject gave answer to the question. Maximum number of topomaps = 7500 as 30 seconds x 250 sampling rate = 7500.

### 3.5. Feature Extraction

The most significant problem in identifying the brain state corresponding to true and false answers of the questions is to extract discriminative features from the related topomaps, which represent the relevant brain state. In this paper, we propose a simple and effective feature extraction technique based on the eigenvalues of the covariance matrix.

First, we form the matrix of all topomaps related to the same question by vectoring each topomap into a column vector, then mean vector is calculated and subtracted from all topomaps vectors. Finally, the covariance matrix is formed and its eigenvalues are calculated.

Specifically, let $t_1, t_2, …, t_n$ be the topomaps related to the question Q of the subject S. First, extract RGB channels from topomaps and compute eigenvalues from each channel independently. After converting each channel into a vector, the matrix A is formed where

$A = [t_1 \quad t_2 \quad … \quad t_n]$, $t_i$ (i =1… n) being a vector of dimension $d$ where $d$ is the number of features.

Next, the mean vector $m$ of all $t_i$ (i =1... n) is computed, the data is centralized by subtracting $m$ from each $t_i$ i.e. $\varphi_i = t_i - m$ and a new matrix $\varphi = [\varphi_1, \varphi_2, \ldots, \varphi_n]$ is formed. Using $\varphi$, the covariance matrix C is calculated i.e. $C = \varphi \varphi^T$. The dimension of C is $d \times d$ and $d \gg n$, where $n$ is the number of topomaps. As the dimension of $\varphi^T \varphi$ is $n \times n$ and largest $n$ eigenvalues of $\varphi^T\varphi$ and $\varphi\varphi^T$ are the same, so for computational efficiency, we calculate the eigenvalues $\lambda_1, \lambda_2, \ldots, \lambda_n$ of $B = \varphi^T \varphi$.

The number of eigenvalues depends on the number of topomaps associated with a question. We sort eigenvalues in descending order, select $k$ (a fixed number for all questions such that $k<100$) largest eigenvalues and use them as features to represent the brain state corresponding to a question. These features clearly discriminate the samples corresponding to two classes as shown in Fig. 2, 3.

### 3.6. Classification

Memory recall and learning processes incorporates two classes of data, i.e., incorrect and correct answers.

For this type of problem, many learning models can be used, which includes but not limited to, K- Nearest Neighbor (KNN), support vector machines (SVM), decision trees, artificial neural networks, etc. Out of these methods; KNN is more suitable with the eigenvalues features [nuclear norm] and also support vector machines are regarded as the state-of-the-art classification models which can achieve outstanding accuracies. SVM is a linear classifier which achieves the maximum margin between the classes of data and can work for the very few training samples and at the same time works for classifying non-linear data as well by taking the data to higher-dimensional space using kernel trick. Various kinds of kernels can be used with SVM like radial basis function (RBF) and polynomial. As RBF kernel is known to give good results, we choose it for the classification purposes [30-33]. The parameters of the kernel ( c and gamma ) are learned using grid-search method, and the well-known libsvm [34] library is used in the implementation of SVM. We split the data into 50:50 training:testing base.

## 4. Results and Discussion

In this section, first, we present the time used by subjects to solve the questions in STM and LTM. Then, we present the results obtained using the proposed eigenvalues based feature extraction technique and discuss them. After that, we show a comparison with the previous techniques applied to the same data [3, 25, 26].

### 4.1 Time:

The mean time used by the subjects to solve the twenty questions, either correctly or incorrectly, in both cases STM and LTM for both 2D and 3D is shown in tables 1 and 2.

As we can see from the tables, 3D time is less than 2D in both STM and LTM for correct and incorrect answers. Furthermore, STM is less than LTM for both correct and incorrect answers. Furthermore, correct is less than incorrect for both STM and LTM. Note that time is represented in seconds.

Table1 : Time for 2D questions

| 2D  | Correct  | Incorrect |
|-----|----------|-----------|
| STM | 8.8±4.6  | 11.6±5.6  |
| LTM | 10.8±4.7 | 12.6±6.1  |

Table2 : Time for 3D questions

| 3D  | Correct | Incorrect |
|-----|---------|-----------|
| STM | 8.5±4.2 | 10.9±5.9  |
| LTM | 9.6±4.2 | 12.2±6.5  |

### 4.2 Eigenvalue Features Classification Results

Applying the proposed feature extraction technique, we implement two systems: one each for 2D and 3D in both STM and LTM.

We select a sample of 200 questions; 100 questions with correct answers and 100 questions with incorrect answers. We limit our selection to 200 questions because we only have 120 correct questions in case of 3D in LTM, and so to avoid imbalanced problem in classification. The selection is done based on the time, i.e. the correct answer with the less time and the incorrect answers with the more time. This selection is new in this research and not like the previous researches [3, 25, 26] where the selection of correct answer from the subjects who gave more correct answers and the incorrect answers from the one with more incorrect answers. We think this selection criteria is best suited to reflect the actual brain states while answering the question. By this selection, all the 66 subjects are participated in the data, and the number of questions selected from each subject is depends on the time.

For evaluation, we used 10-fold cross-validation and percentage accuracy, which is commonly used for performance measure of a pattern recognition system. The results have been reported as average accuracy along with standard deviation (accuracy±std) over 10 folds.

### 4.2.1 STM

To see the effects of choosing eigenvalues as features we select the highest two eigenvalues ($\lambda 1$, $\lambda 2$) and plot them according to their class (Correct and Incorrect) in each channel R, G and B as shown in the following figures:

We start with the 2D questions (samples):

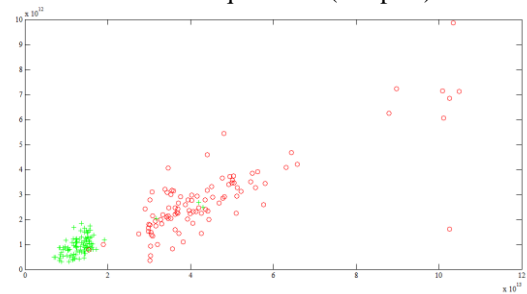

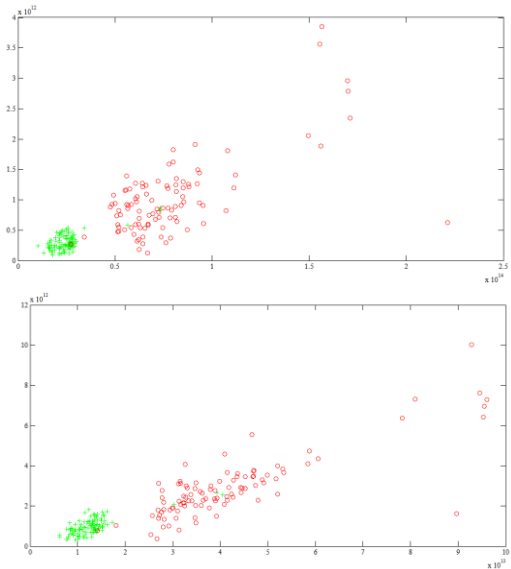

Figure 2 (a,b,c): The visualization of Correct (green) and Incorrect (red) answer instances in the feature space with two features (the highest two eigenvalues) in the Red (a.), Green (b.) and Blue (c.) Channel-2D

It is clear from the figures that these two features clearly separate the classes with little outliers only. In addition, we notice that all channels are nearly having the same margin characteristics.

Then we see the 3D questions (samples) and repeat the same highest two eigenvalues selection:

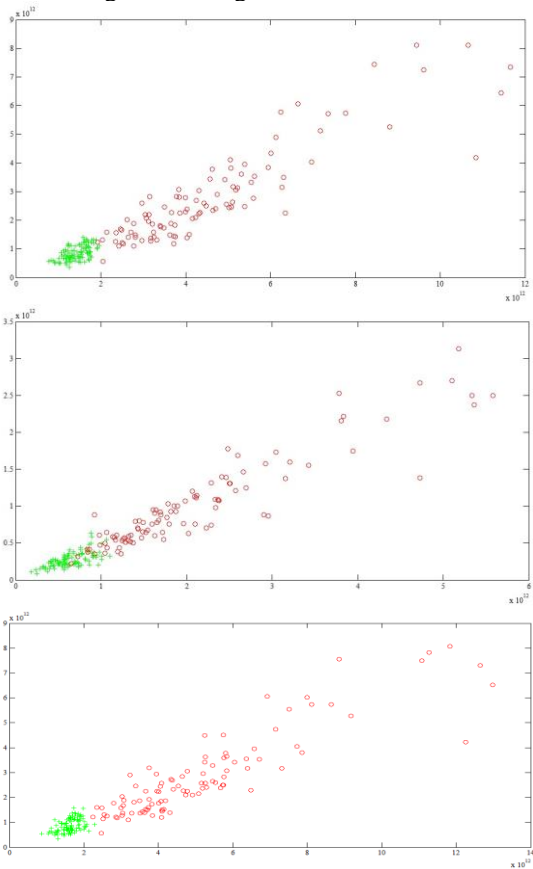

Figure 3(a,b,c): The visualization of Correct (green) and Incorrect (red) answer instances in the feature space with two features (the highest two eigenvalues) in the Red (a.), Green (b.) and Blue (c.) Channel -3D

We notice that the Green channel is the worst one of the highest two eigenvalues in the 3D case. In addition, the margin of all the three channel in 3D case is not like those in 2D case.

Now we plot the accuracies of the highest 100 eigenvalues for 2D using KNN:

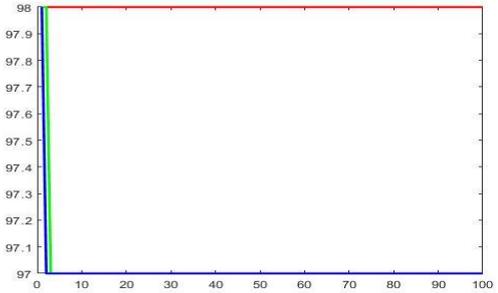

Figure 4: KNN accuracies of the highest 100 eigenvalues for R (red), G (green) and B (blue) in 2D

We notice that R channel gave the best and it is consistent for all the 100 eigenvalues. Accuracy is 98%.

Now we plot the accuracies of the highest 100 eigenvalues for 2D using SVM:

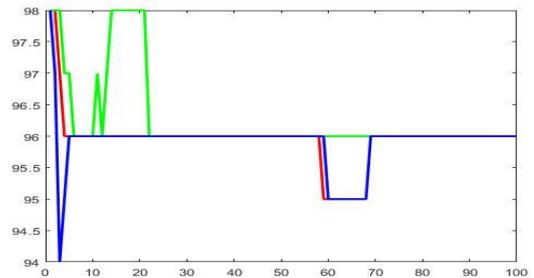

Figure 5: SVM accuracies of the highest 100 eigenvalues for R (red), G (green) and B (blue) in 2D

We notice that R, G, and B channels gave the best accuracy with the highest 1, 2 and 3 eigenvalues. Accuracy is 98%. Then when we increase the number of eigenvalues, accuracy decreased and increased.

Now we plot the accuracies of the highest 100 eigenvalues for 3D using KNN:

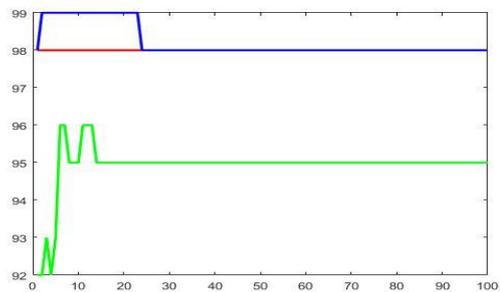

Figure 6: KNN accuracies of the highest 100 eigenvalues for R (red), G (green) and B (blue) in 3D

We notice that R and B channels gave the best accuracies (98-99%). It is clear that G is the worst channel in this case.

Now we plot the accuracies of the highest 100 eigenvalues for 3D using SVM:

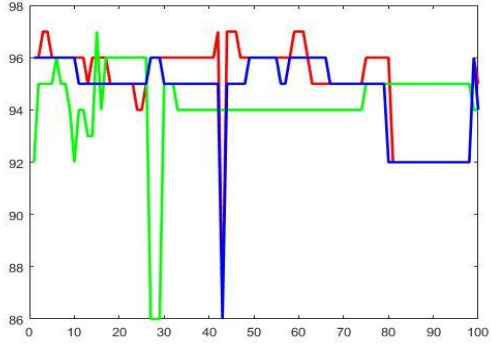

Figure 7: SVM accuracies of the highest 100 eigenvalues for R (red), G (green) and B (blue) in 3D

The best accuracies is given by the highest eigenvalues for R channel. Other's accuracies are goes up and down and it reaches 86% for G channel with highest 28 and 29 Eigen.

Now we choose the highest three eigenvalues and run each system (2D and 3D) with SVM and KNN, 10 times with different randomization at each run.

Majority vote means the majority of the three classifiers of the red, green and blue channels. RGB is obtained by combining the Eigen features of all the three channels R, G and B together.

Table 3: SVM accuracy of the highest 3 Eigen values in 2D

| SVM Accuracy | | | | |
|---|---|---|---|---|
| Majority Vote | R | G | B | RGB |
| 96.5±2.5 | 96.2±2.5 | 97.1±3.1 | 95.9±2.8 | 95.8±3.7 |

Table 4: SVM accuracy of the highest 3 Eigen values in 3D

| SVM Accuracy | | | | |
|---|---|---|---|---|
| Majority Vote | R | G | B | RGB |
| 98±1.5 | 98±1.5 | 95.3±5.4 | 98±1.5 | 95.3±5.4 |

Table 5: KNN accuracy of the highest 3 Eigen values in 2D

| KNN Accuracy | | | | |
|---|---|---|---|---|
| K | R | G | B | RGB |
| 1 | 95.5±1.9 | 95.8±2.2 | 94.6±1.9 | 94.3±2.3 |
| 3 | 98.1±0.9 | 97.3±1.4 | 97.9±0.9 | 97.6±0.7 |
| 5 | 98.1±0.9 | 98.1±0.9 | 98.1±0.9 | 98.1±0.9 |
| 7 | 98.1±0.9 | 98.1±0.9 | 98.1±0.9 | 98.1±0.9 |

Table 6: KNN accuracy of the highest 3 Eigen values in 3D

| KNN Accuracy | | | | |
|---|---|---|---|---|
| K | R | G | B | RGB |
| 1 | 98.3±0.6 | 91.7±1.5 | 99±0 | 98.3±0.6 |
| 3 | 98.3±0.6 | 93.3±0.6 | 99±0 | 98.3±0.6 |
| 5 | 98.3±0.6 | 93.7±0.6 | 98.7±0.6 | 98±1 |
| 7 | 98.3±0.6 | 93.7±0.6 | 98.7±0.6 | 98±1 |

### 4.2.2 LTM:

Now we will present the results in case of LTM in the same manner as we presented those for STM in the previous section.

First, we will start with the effects of choosing two eigenvalues as features and plot them according to their class (Correct and Incorrect) in each channel R, G and B in 2D case as shown in the following figures:

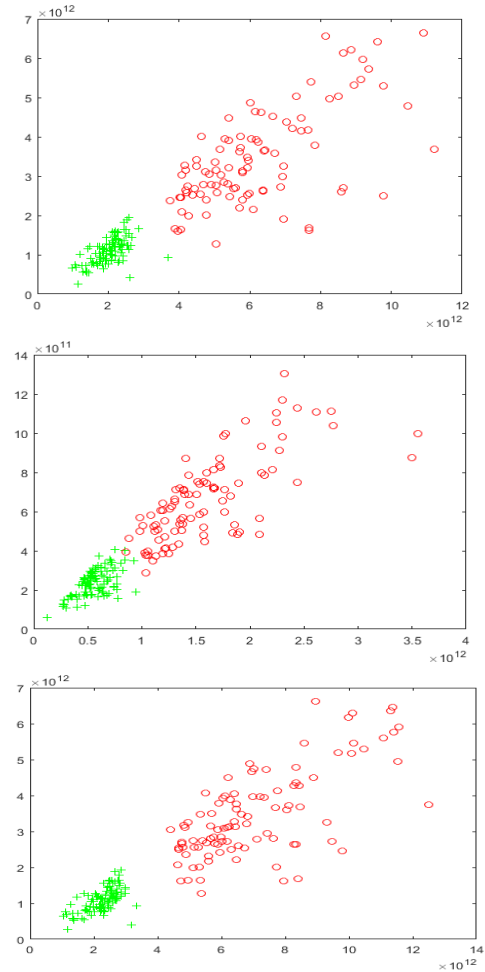

Figure 8(a,b,c): The visualization of Correct (green) and Incorrect (red) answer instances in the feature space with two features (the highest two eigenvalues) in the Red (a.), Green (b.) and Blue (c.) Channel-2D

The same observation obtained in STM can be seen from the last three figures that these two features clearly separate the classes with little outliers only. In addition, LTM Eigen values are different from those of STM.

Now we will show the results for the 3D questions (samples) and repeat the same highest two Eigen values selection:

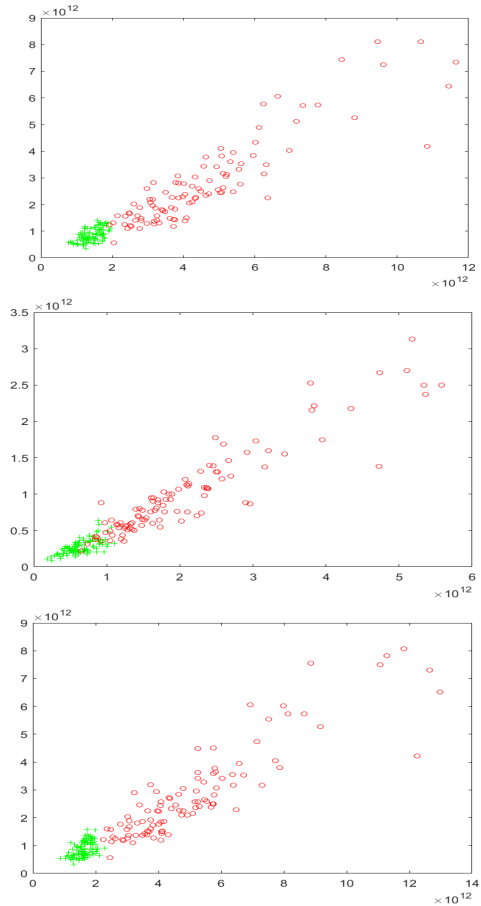

Figure 9(a,b,c): The visualization of Correct (green) and Incorrect (red) answer instances in the feature space with two features (the highest two eigenvalues) in the Red (a.), Green (b.) and Blue (c.) Channel-3D

Now we plot the accuracies of the highest 100 eigenvalues for 2D using KNN:

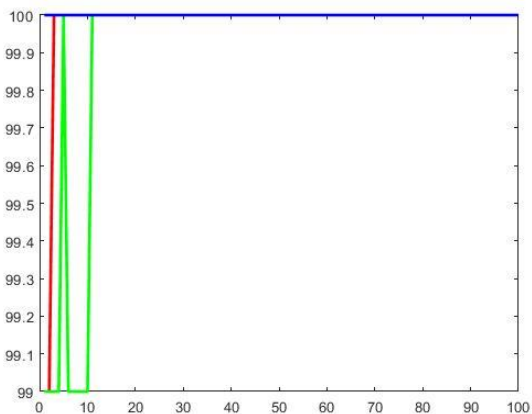

Figure 10: KNN accuracies of the highest 100 Eigen values for R (red), G (green) and B (blue) in 2D

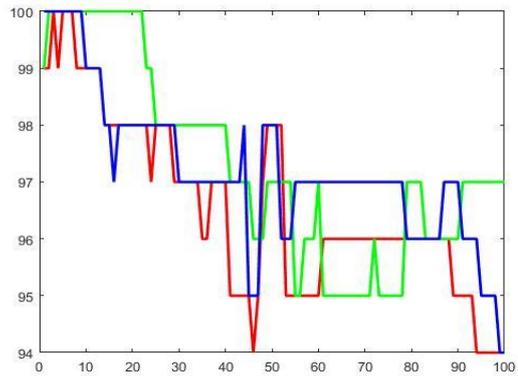

Figure 11: SVM accuracies of the highest 100 Eigen values for R (red), G (green) and B (blue) in 2D

We notice that R, G, and B channels gave the best accuracy with the highest 1, 2 and 3 Eigen. Accuracy is 98%. Then when increase the number of Eigen accuracy decreased and increased.

Now we plot the accuracies of the highest 100 Eigen values for 3D using KNN:

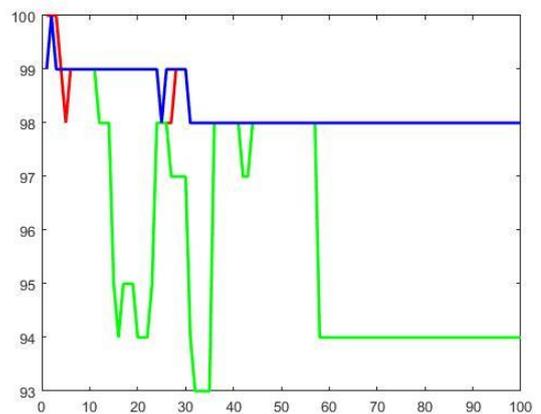

Figure 12: KNN accuracies of the highest 100 Eigen values for R (red), G (green) and B (blue) in 3D

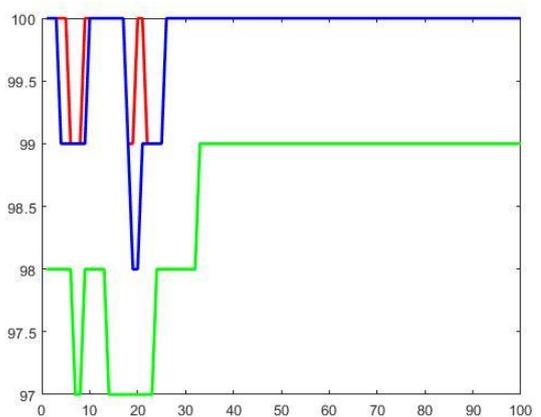

Figure 13: SVM accuracies of the highest 100 Eigen values for R (red), G (green) and B (blue) in 3D

The best accuracies are given by the highest 3 Eigen values for R channel.

Now we choose the highest three Eigen values and run each system (2D and 3D) with SVM and KNN, ten times with different randomization at each run.

Table 7: SVM accuracy of the highest 3 Eigen values in 2D

| SVM Accuracy | | | | |
|---|---|---|---|---|
| Majority Vote | R | G | B | RGB |
| 98.6±1.4 | 97.4±1.5 | 97.7±1.5 | 98±1.6 | 98.2±1.8 |

Table 8: SVM accuracy of the highest 3 Eigen values in 3D

| SVM Accuracy | | | | |
|---|---|---|---|---|
| Majority Vote | R | G | B | RGB |
| 97.8±1.9 | 99.2±1.1 | 97±2.1 | 97.8±1.8 | 98.3±1.2 |

Table 9: KNN accuracy of the highest 3 Eigen values in 2D

| KNN Accuracy | | | | |
|---|---|---|---|---|
| K | R | G | B | RGB |
| 1 | 99.8±0.4 | 98.8±0.6 | 100 | 100 |
| 3 | 100 | 99.1±0.7 | 100 | 100 |
| 5 | 99.9±0.3 | 99.4±0.6 | 100 | 100 |
| 7 | 100 | 99.1±0.6 | 100 | 100 |

Table 10: KNN accuracy of the highest 3 Eigen values in 2D

| KNN Accuracy | | | | |
|---|---|---|---|---|
| K | R | G | B | RGB |
| 1 | 99.5±0.5 | 99.3±0.9 | 99.7±0.5 | 99.5±0.5 |
| 3 | 99.4±0.5 | 99.6±0.5 | 100 | 99.7±0.5 |
| 5 | 99.4±0.5 | 99.4±0.7 | 99.9±0.3 | 99.7±0.5 |
| 7 | 99.4±0.5 | 99±0.8 | 99.9±0.3 | 99.7±0.5 |

### 4.3 Comparison with previous methods:

The comparison of the work in this paper with the techniques used in the previous researches [3, 25, 26] can be shown in terms of the number of features and the number of topomaps. This research gave us the best accuracy using up to three features only compared to 5120 features in the past researches. In addition, this research used all the topomaps of each question without any selection as in the previous researches, but this led to increasing in the time of the execution.

### 4.4 Which is better 2D or 3D?

Though results presented and discussed in the previous sections indicate that there is no difference between 2D and 3D educational content, it needs to be verified using statistical significance test.

To find out the significant difference among 2D and 3D educational contents for Short-Term Memory (STM) and Long-Term Memory (LTM), we need enough samples of accuracy values. For this purpose, we run each system 3 times with 10-fold cross validation using the best parameters and every time randomizing the datasets. This gave us thirty accuracies for each system. An independent t-test was utilized by using the SPSS software based on this assumption that normality is acceptable. No significant difference was shown in the values for 2D and 3D, $p = 0.5$. Therefore, we can conclude that no statistically significant difference exists between 3D and 2D multimedia educational content in terms of learning and memory recall in STM and LTM.

## Conclusions

In this research work, we presented and discussed the results on the impacts of 3D and 2D multimedia educational contents on memory retention and recall, for STM and LTM utilizing EEG signals, which reflect the states of the brain. This problem is modeled as a classification problem. We proposed an eigenvalues based feature extraction technique to extract features from topomaps. Features are passed to Support Vector Machine (SVM) and KNN classifiers to predict brain states corresponding to correct/incorrect answers. The results of this research work indicate that 3D educational content gave 99% classification accuracy as compared to 98% given by 2D educational content in STM, whereas in case of LTM, 3D gave 99.5% classification accuracy and 2D showed 98.5% accuracy. Statistical analysis of the outputs shows no significant difference exists between 3D and 2D multimedia educational content in STM and LTM on learning, memory retention and recall processes.

## Compliance with ethical standards

**Conflicts of interest** The authors have no conflicts of interest to declare.

**Ethical approval** All procedures performed in studies involving human participants were in accordance with the ethical standards of the institutional and/or national research committee and with the 1964 Helsinki declaration and its later amendments or comparable ethical standards.

# References

[1] 1. Mayer, R.E. and R. Moreno, *A cognitive theory of multimedia learning: Implications for design principles.* Journal of Educational Psychology, 1998. **91**(2): p. 358-368.

[2] 2. Niedermeyer, E. and F.L. da Silva, *Electroencephalography: basic principles, clinical applications, and related fields* 2005: Lippincott Williams & Wilkins.

[3] 3. Bamatraf, S., et al., *A System for True and False Memory Prediction Based on 2D and 3D Educational Contents and EEG Brain Signals.* Computational Intelligence and Neuroscience, 2015. **2016**.

[4] 4. Zhang, F., et al., *Nuclear Norm-Based 2-DPCA for Extracting Features From Images.* Neural Networks and Learning Systems, IEEE Transactions on, 2015. **26**(10): p. 2247-2260.

[5] 5. Cockburn, A. and B. McKenzie, *Evaluating spatial memory in two and three dimensions.* International Journal of Human-Computer Studies, 2004. **61**(3): p. 359-373.

[6] 6. Price, A. and H.-S. Lee, *The effect of two-dimensional and stereoscopic presentation on middle school students' performance of spatial cognition tasks.* Journal of Science Education and Technology, 2010. **19**(1): p. 90-103.

[7] 7. Tavanti, M. and M. Lind. *2D vs 3D, implications on spatial memory.* in *Information Visualization, 2001. INFOVIS 2001. IEEE Symposium on*. 2001. IEEE.

[8] 8. Wang, H.-C., C.-Y. Chang, and T.-Y. Li, *The comparative efficacy of 2D-versus 3D-based media design for influencing spatial visualization skills.* Computers in Human Behavior, 2007. **23**(4): p. 1943-1957.

[9] 9. Mukai, A., et al., *Effects of Stereoscopic 3D Contents on the Process of Learning to Build a Handmade PC.* Knowledge Management & E-Learning: An International Journal (KM&EL), 2011. **3**(3): p. 491-506.

[10] 10. Carrier, L.M., et al., *Pathways for Learning from 3D Technology.* International Journal of Environmental & Science Education, 2012. **7**(1).

[11] 11. Huk, T. and C. Floto. *Computer-animations in education: the impact of graphical quality (3D/2D) and signals.* in *World Conference on E-Learning in Corporate, Government, Healthcare, and Higher Education*. 2003.

[12] 12. Rias, R.M. and W. Khadijah Yusof. *Animation and prior knowledge in a multimedia application: A case study on undergraduate computer science students in learning.* in *Digital Information and Communication Technology and it's Applications (DICTAP), 2012 Second International Conference on*. 2012. IEEE.

[13] 13. Perez, F., S. Vanegas, and P. Cortes. *Evaluation of learning processes applying 3D models for constructive processes from solutions to problems.* in *Education Technology and Computer (ICETC), 2010 2nd International Conference on*. 2010. IEEE.

[14] 14. Amin, H.U., et al. *Brain activation during cognitive tasks: An overview of EEG and fMRI studies.* in *Biomedical Engineering and Sciences (IECBES), 2012 IEEE EMBS Conference on*. 2012. IEEE.

[15] 15. Malik, A.S., et al. *Investigating brain activation with respect to playing video games on large screens.* in *Intelligent and Advanced Systems (ICIAS), 2012 4th International Conference on*. 2012. IEEE.

[16] 16. Malik, A.S., et al. *Disparity in brain dynamics for video games played on small and large displays.* in *Humanities, Science and Engineering Research (SHUSER), 2012 IEEE Symposium on*. 2012. IEEE.

[17] 17. Amin, H.U., et al. *Brain Behavior in Learning and Memory Recall Process: A High-Resolution EEG Analysis.* in *The 15th International Conference on Biomedical Engineering*. 2014. Springer.

[18] 18. Babiloni, C., et al., *Human cortical responses during one-bit short-term memory. A high-resolution EEG study on delayed choice reaction time tasks.* Clinical Neurophysiology, 2004. **115**(1): p. 161-170.

[19] 19. CHEN, X.-l., B. YAO, and D.-y. WU, *Application of EEG Non-linear Analysis in Vision Memory Study.* Chinese Journal of Rehabilitation Theory and Practice, 2006. **6**: p. 012.

[20] 20. Klimesch, W., *EEG alpha and theta oscillations reflect cognitive and memory performance: a review and analysis.* Brain research reviews, 1999. **29**(2): p. 169-195.

[21] 21. Coffman, B.A., V.P. Clark, and R. Parasuraman, *Battery powered thought: enhancement of attention, learning, and memory in healthy adults using transcranial direct current stimulation.* NeuroImage, 2014. **85**: p. 895-908.

[22] 22. Gu, B.-M., H. van Rijn, and W.H. Meck, *Oscillatory multiplexing of neural population codes for interval timing and working memory.* Neuroscience & Biobehavioral Reviews, 2015. **48**: p. 160-185.

[23] 23. Hsieh, L.-T. and C. Ranganath, *Frontal midline theta oscillations during working memory maintenance and episodic encoding and retrieval.* NeuroImage, 2014. **85**: p. 721-729.

[24] 24. Johnson, E.L. and R.T. Knight, *Intracranial recordings and human memory.* Current opinion in neurobiology, 2015. **31**: p. 18-25.

[25] 25. Bamatraf, S., et al. *Studying the Effects of 2D and 3D Educational Contents on Memory Recall Using EEG Signals, PCA and Statistical Features.* in *Artificial Intelligence, Modelling and Simulation (AIMS), 2014 2nd International Conference on*. 2014. IEEE.

[26] 26. Bamatraf, S., et al. *A System based on 3D and 2D Educational Contents for True and False Memory Prediction using EEG Signals.* in *7th Annual International IEEE EMBS Conference on Neural Engineering*. 2015. Montpellier, France: IEEE.

[27] 27. Geodesics, E., *Geodesic Sensor Net Technical Manual.* Eugene: Electrical geodesics, 2007.

[28] 28. Gratton, G., M.G. Coles, and E. Donchin, *A new method for off-line removal of ocular artifact.* Electroencephalography and clinical neurophysiology, 1983. **55**(4): p. 468-484.

[29] 29. Delorme, A. and S. Makeig, *EEGLAB: an open source toolbox for analysis of single-trial EEG dynamics including independent component analysis.* Journal of neuroscience methods, 2004. **134**(1): p. 9-21.

[30] 30. Joachims, T., *Text categorization with support vector machines: Learning with many relevant features* 1998: Springer.

[31] 31. Cortes, C. and V. Vapnik, *Support-vector networks.* Machine learning, 1995. **20**(3): p. 273-297.

[32] 32. Chapelle, O., P. Haffner, and V.N. Vapnik, *Support vector machines for histogram-based image*

*classification.* Neural Networks, IEEE Transactions on, 1999. **10**(5): p. 1055-1064.
[33] 33. Hsu, C.-W., C.-C. Chang, and C.-J. Lin, *A practical guide to support vector classification.* 2003.
[34] 34. Chang, C.-C. and C.-J. Lin, *LIBSVM: a library for support vector machines.* ACM Transactions on Intelligent Systems and Technology (TIST), 2011. **2**(3): p. 27.